\documentclass[10pt,twocolumn,letterpaper]{article}

\usepackage[pagenumbers]{posecraft} %

\usepackage[dvipsnames]{xcolor}

\IfFileExists{orcidlink.sty}{%
  \usepackage{orcidlink}%
}{%
  \newcommand{\orcidlink}[1]{}%
}

\usepackage{multirow} 

\def\paperID{29620} %
\def\confName{arXiv\xspace}
\def\confYear{2026\xspace}

\usepackage{xcolor}
\usepackage{colortbl}
\usepackage{booktabs}
\definecolor{lightred}{rgb}{1,0.55,0.55}
\definecolor{lightorange}{rgb}{1,0.8,0.8}
\definecolor{lightyellow}{rgb}{1,0.99,0.8}

\title{PoseCraft: Tokenized 3D Body Landmark and Camera Conditioning for Photorealistic Human Image Synthesis}

\author{
Zhilin Guo \orcidlink{0000-0002-7660-3102}~$^{1}$ \quad
Jing Yang \orcidlink{0000-0002-8794-4842}~$^{1}$ \quad
Kyle Fogarty \orcidlink{0000-0002-1888-4006
}~$^{1}$ \quad
Jingyi Wan \orcidlink{0009-0002-0623-8637}~$^{1}$ \quad
Boqiao Zhang \orcidlink{0009-0000-0745-2438}~$^{1}$ \\
Tianhao Wu \orcidlink{0000-0002-3807-5839}~$^{1}$ \quad
Weihao Xia \orcidlink{0000-0003-0087-3525
}~$^{1}$ \quad
Chenliang Zhou \orcidlink{0009-0004-4687-6945
}~$^{1}$ \quad
Sakar Khattar \orcidlink{0009-0008-6707-959X}~$^{2}$ \\
Fangcheng Zhong \orcidlink{0000-0001-5964-5282
}~$^{1\dagger}$ \quad
Cristina Nader Vasconcelos \orcidlink{0000-0003-2112-4806
}~$^{2\dagger}$ \quad
Cengiz Oztireli \orcidlink{0000-0002-4700-2236}~$^{1,2\dagger}$ \\
{\small $^{1}$University of Cambridge \quad
$^{2}$Google}
}

\begin{document}
\maketitle
\begingroup
\renewcommand\thefootnote{\fnsymbol{footnote}}
\footnotetext[2]{Corresponding author.}
\footnotetext[0]{This work was supported by a UKRI Future Leaders Fellowship [grant number G127262].}
\endgroup

\begin{abstract}

Digitizing humans and synthesizing photorealistic avatars with explicit 3D pose and camera controls are central to VR, telepresence, and entertainment.
Existing skinning-based workflows require laborious manual rigging or template-based fittings, while neural volumetric methods rely on canonical templates and re-optimization for each unseen pose.
We present PoseCraft, a diffusion framework built around tokenized 3D interface: instead of relying only on rasterized geometry as 2D control images, we encode sparse 3D landmarks and camera extrinsics as discrete conditioning tokens and inject them into diffusion via cross-attention. 
Our approach preserves 3D semantics by avoiding 2D re-projection ambiguity under large pose and viewpoint changes, and produces photorealistic imagery that faithfully captures identity and appearance.
To train and evaluate at scale, we also implement GenHumanRF, a data generation workflow that renders diverse supervision from volumetric reconstructions.
Our experiments show that PoseCraft achieves significant perceptual quality improvement over diffusion-centric methods, and attains better or comparable metrics to latest volumetric rendering SOTA while better preserving fabric and hair details.

\end{abstract}

\section{Introduction}
\begin{figure*}[t]
  \centering
  \includegraphics[width=1.0\textwidth]{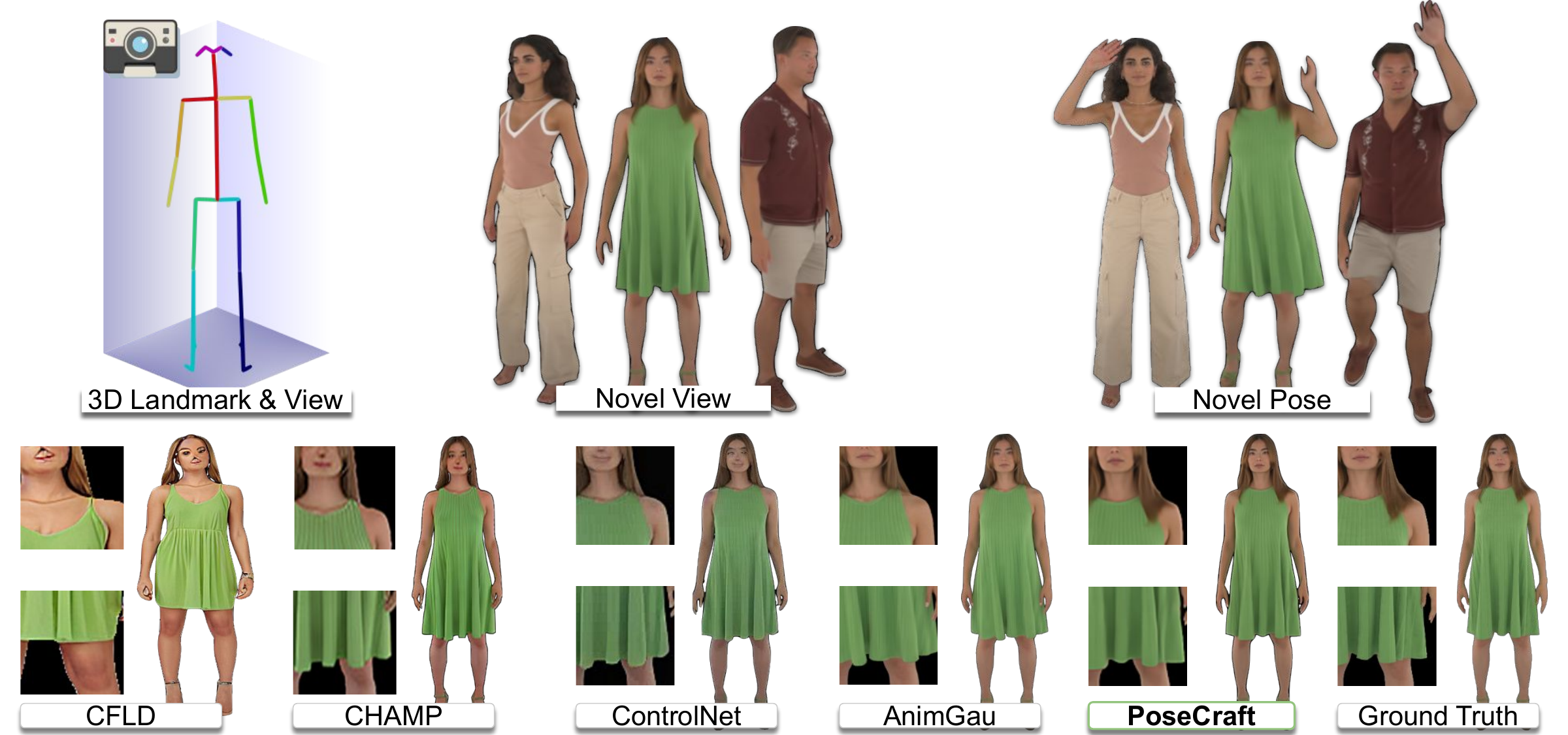}
  \caption{
     PoseCraft uses tokenized 3D landmarks and camera parameters to synthesize photorealistic humans. This explicit 3D control delivers sharp silhouettes, preserves high-frequency details, and ensures structural coherence across novel viewpoints.
  }
  \label{fig:teaser-image}
\end{figure*}

From virtual reality and gaming to telepresence and digital content creation, digitizing human then synthesizing photorealistic avatar images is a fundamental task underpinning a wide range of applications.
Practical deployment demands avatars that are both visually faithful and precisely controllable in 3D, as creators must set pose and camera explicitly while system preserves identity and appearance details across novel stances and views.
But achieving this goal is challenging: it requires disentangling pose and viewpoint from appearance, avoiding re-projection ambiguity, and maintaining geometric coherence under articulation and camera changes.
Additionally, subject-specific capture is often limited, so solutions must work with limited pose distribution and sparse signals, ideally without fixed template or scene-specific optimization.

Existing computer graphics techniques manually sculpt and rigs 3D characters to control shape, pose, and appearance, but are often laborious and resource intensive.
Parametric body models such as SMPL and SMPL-X~\cite{loper2015smpl, pavlakos2019smplx} streamline body rigging by utilising template-based blend shapes, but struggle to represent complex body geometry and appearance details~\cite{prokudin2021smplpix, corona2021smplicit, ma2022neural, pmlr-v133-madadi21a}.
Complementary to mesh-based pipelines, neural volumetric rendering has shown strong view synthesis performance~\cite{nerf, kerbl3Dgaussians}.
Recent avatar-oriented adaptations use either injection of pose or skeleton priors into implicit radiance fields~\cite{li2023posevocab, zheng2022slrf, wang2022arah, li2022tava}, or optimising point-based canonical body template~\cite{li2024animatablegaussians} to synthesize novel poses.
But these approaches either bias towards low-frequency structure and produce blurry details, or depend on fixed template that struggles to create realistic appearance alterations under novel poses.

From the rapid progress of novel image generators comes a plethora of 2D image-based identity retargeting and pose transfer models spanning classic GANs and modern diffusion backbones that deliver striking image quality~\cite{zhu2019patn, goodfellow2020gan, men2020adgan, ren2020gfla, ho2020ddpm, song2020ddim, rombach2022ldm, Zhang2021pise, lv2021spgnet, zhang2022dptn, ren2022nted, zhou2022casd, bhunia2022pidm, han2023pocold, lu2024cfld}.
Recent diffusion pipelines 
build on latent or pixel-space denoising
by adapting strong text-to-image backbones,
and inject structure via 2D conditioning such as skeleton heatmaps, dense correspondences, silhouettes, or depth maps ~\cite{bhunia2022pidm, lu2024cfld, ju2023humansd, mou2024t2i, zhang2023controlnet, liu2024hyperhuman}.
However, 2D conditioning leaves pose and depth ambiguous or occluded under large articulations and viewpoint changes, yielding viewpoint-inconsistent hallucinations and drifts around limbs and loose garments.

We therefore aim for a formulation that easily digitizes a human subject and exposes 3D control while producing identity-preserving, viewpoint-consistent imagery in a template-free setting.
In this paper, we present \emph{PoseCraft}, a photorealistic image-diffusion pipeline designed for controllable human digitization.
We design \emph{PoseCraft} around a tokenized 3D interface: instead of relying exclusively on rasterizing geometry into control images, we encode sparse 3D landmarks and camera extrinsics as discrete conditioning tokens and inject them into a diffusion backbone via cross-attention.
This 3D tokenization and interface keeps 3D pose and view explicit and unambiguous.

Concretely, our \emph{RigCraft} supplies template-free, temporally stable 3D landmarks from multi-view detections; \emph{PoseCraft} encodes and consumes these as tokens to constrain coarse geometry while the denoiser focuses on high-frequency appearance, yielding identity-preserving and viewpoint-consistent images across novel poses and viewpoints.
The result is a high-fidelity human digitization pipeline without mesh rigging or per-scene optimization.

In summary, our contributions are: 
\begin{itemize}
\item \textbf{RigCraft}: Uncertainty-aware multi-view fusion with temporal smoothing for stable 3D landmarks,
\item \textbf{PoseCraft}: 3D control tokenizer and image-diffusion for photorealistic synthesis under pose/view control,
\item \textbf{GenHumanRF}: A volumetric rendering pipeline for large-scale, 3D-supervised diffusion training.
\end{itemize}

\section{Related Works}

\begin{figure*}[t]
  \centering
  \includegraphics[width=0.95\textwidth]{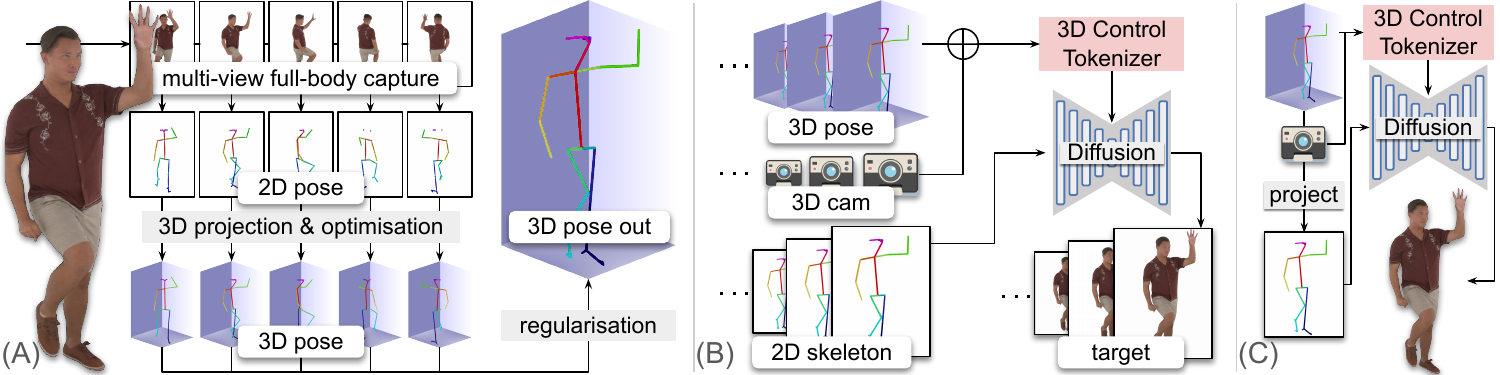}
  \caption{
    \textbf{Pipeline overview} of (A) RigCraft and (B, C) PoseCraft.
    (A) RigCraft performs multi-view fusion to generate temporally stable 3D landmarks from 2D poses.
    (B) During training, PoseCraft learns to denoise latent representations conditioned on tokenized 3D landmarks and camera parameters.
    (C) At inference, PoseCraft synthesizes a photorealistic image from a target 3D pose and camera view.
  }
  \label{fig:pipeline-overview}
\end{figure*}

\subsection{3D Human Capture \& Rendering}

For human body rendering, existing 3D graphics pipelines employ skinned polygon meshes attached to articulated skeletal rigs where artists manually sculpt and rig meshes so poses stays visually plausible across movement.
However, this process is labour-intensive and often produces visual artifacts under unseen poses.
Consequently, researchers have explored data-driven solutions to alleviate manual rigging effort and better capture anatomical details.
Notable examples include Skinned Multi Person Linear Model (SMPL)~\cite{loper2015smpl} and its extension SMPL-eXpressive (SMPL-X)~\cite{pavlakos2019smplx}, which are learned parametric body models combining linear blend skinning with pose-corrective blend shapes to jointly model body shape and pose.
Yet despite more accessible pose modelling and manipulation, mesh-based renderings produced by SMPL often struggle to recreate fine-grained surface geometry and texture such as wrinkled fabrics or hair, and remain sensitive to out-of-distribution poses.
Our approach bypasses complex manual rigging but instead fuses 3D skeletal data and camera extrinsics as conditioning tokens for diffusion, allowing diffusion to learn accurate pose and high-fidelity texture details. 

On the other hand, advances in neural volumetric rendering such as NeRF~\cite{nerf} and 3DGS~\cite{kerbl3Dgaussians} have given rise to volumetric 3D human representations.
Recent adaptations of NeRF introduce novel approaches for creating human avatars~\cite{li2023posevocab, zheng2022slrf, wang2022arah, li2022tava} that inject pose or skeleton priors into implicit radiance fields, whereas AnimatableGaussians~\cite{li2024animatablegaussians} leverages 3DGS and per-pose canonical template optimization.
Nonetheless, these works either rely on MLP backbones biased towards low-frequency structures~\cite{tancik2020fourfeat} that appear blurry, or depend on manipulating a canonical point template which cannot fully capture pose and appearance variations observed in real-world environment.
In contrast, our approach uses sparse 3D landmarks to optimize image diffusion model that is high-fidelity in nature, creating more realistic human images.

\subsection{Human-Pose Guided Diffusion}

Building on the successes of unconditional image diffusion~\cite{ho2020ddpm, song2020ddim, rombach2022ldm} and text-guided or image-guided diffusion~\cite{rombach2022ldm, xia2021tedigan, zhang2023controlnet}, pose-guided diffusion aims to denoise latent representations under pose constraints to produce controlled human images.
Among these approaches, PIDM~\cite{bhunia2022pidm} enriches a diffusion network with skeleton guidance, whereas PoCoLD~\cite{han2023pocold} uses latent features of both source and target poses to align appearance within a compressed space.
More recently, HumanSD~\cite{ju2023humansd} introduces native skeleton control that fine-tunes Stable Diffusion with heatmap-guided denoising loss;
Coarse-to-Fine Latent Diffusion (CFLD)~\cite{lu2024cfld} learns a coarse semantic prompt and later adding fine details;
HyperHuman~\cite{liu2024hyperhuman} presents Latent Structural Diffusion that jointly denoises RGB, depth, and surface normals to couple appearance with multi-granularity human structure;
and T2I-Adapter~\cite{mou2024t2i} proposes lightweight adapter that attach to a frozen text-to-image diffusion model to align external spatial conditions including pose keypoints.
However, larger pose variations still triggers image distortions and view-point inconsistencies, and these works continues to rely on 2D control maps (\eg openpose keypoints) rather than explicit 3D representations.
Our work addresses this gap by employing 3D body landmarks and camera extrinsics to condition the diffusion model, enhancing geometric robustness and consistency across viewpoints.

Similarly, some recent video-centric diffusion models also condition on 2D poses despite their emphasis on temporal smoothness.
Among these works, DreamPose\cite{karras2023dreampose} creates fashion videos from 2D human pose sequence, AnimateAnyone~\cite{hu2024animate} attempts arbitrary character animation with appearance preservation, and MimicMotion~\cite{zhang2025mimicmotion} incorporates confidence-aware pose guidance to stabilize longer video clips.
Further developments incorporate more explicit 3D signals or viewpoints, such as Human4DiT~\cite{shao2024human4dit}, a 4D diffusion transformer conditioned on rendered SMPL sequences for human video;
Diffusion-as-Shader~\cite{gu2025diffusion-as-shader}, which leverages mesh renderings as control for 3D-aware generation;
and Champ~\cite{zhu2024champ}, which injects SMPL-based renderings to transfer source identity to target pose.
Despite their optimization for video, these approaches still rely on 3D information visualized to 2D images, leading to ambiguity and inefficiency.
In contrast, our approach performs single-image photorealistic synthesis reinforced by explicit 3D landmarks and camera extrinsics, yielding high image quality and strong view-point consistency.

\subsection{Multi-view 3D Human Pose Estimation}

Multi-view 3D human pose estimation aims to recover temporally and spatially consistent skeletal landmarks from calibrated, synchronized cameras.
Recent advances has matured into a spectrum that spans pure geometry and learned cross-view fusion.
VoxelPose~\cite{tu2020voxelpose} aggregates multi-view image features into a common 3D voxel grid, where human subjects are localized before per-proposal 3D pose regression.
GraphPose~\cite{wu2021graph} decomposes 3D human pose into 3D person localization and 3D pose regression, each refining a coarse-to-fine graph neural network to reduce cubic cost of dense 3D volumes while preserving accuracy.
Instead of cross-view matching or 3D volumes, PlaneSweepPose~\cite{lin2021multi} performs per-joint depth regression using plane-sweep stereo in a coarse-to-fine scheme.
EasyMocap~\cite{shuai2021easymocap} is an open-source toolbox for markerless multi-view capture.
Multi-view Pose Lifter (MPL)~\cite{ghasemzadeh2024mpl} lifts synchronized 2D keypoints to 3D with a transformer-based network trained on synthetic 2D–3D pairs.

In contrast, our pipeline targets a complementary goal - \emph{RigCraft} fuses multi-view 2D detections via non-linear probabilistic triangulation with temporal smoothing to produce stable, temporally coherent 3D landmakrs without volumetric grids or learned cross-view matching. This simple yet effective approach turns 3D landmarks into efficient control signals for our \emph{PoseCraft} diffusion architecture.

\section{Method}
\label{sec:method}

Given a sequence of frames capturing a full‐body human subject in diverse poses, we learn a generative model that produces viewpoint and pose-consistent, high‐fidelity imagery of this subject by conditioning on sparse 3D body landmarks and camera parameters.
Illustrated in ~\cref{fig:pipeline-overview}, our framework comprises two pipelines:
(1)
\emph{RigCraft}, a uncertainty-aware triangulation stage that fuses 2D keypoint detections across calibrated views to yield spatially and temporally coherent 3D body landmarks (\cref{subsec:method:rigcraft});
and
(2)
\emph{PoseCraft}, a latent diffusion model that conditions on tokenized 3D body landmarks and camera view to synthesize photorealistic images (\cref{subsec:method:posecraft}).

\subsection{RigCraft: Extracting 3D Landmarks}
\label{subsec:method:rigcraft}

Following the convention of ActorsHQ~\cite{isik2023humanrf}, we define a \textit{frame} as a single image within a temporal sequence, where the human pose evolves over time. Each frame is captured by multiple calibrated cameras, producing a set of images from different \textit{views}, each offering a distinct spatial perspective of the same pose.

\emph{RigCraft} reconstructs 3D landmarks by fusing 2D joint detections from these multi-view images into spatially and temporally consistent 3D keypoints. For each frame, we begin by applying OpenPose~\cite{Cao2019OpenPose} to detect 2D landmarks in every view. For each landmark, we project its 2D detections into the 3D scene to form rays from each camera. The intersection of these rays provides an estimate of the landmark’s 3D position, ensuring spatial consistency across views. To promote temporal coherence and reduce jitter, we apply temporal regularization across frames, yielding smooth and continuous landmark trajectories. These 3D landmarks, which capture the subject’s underlying kinematics, serve as conditioning input for the diffusion-based motion model described in~\cref{subsec:method:posecraft}.

\subsubsection{Camera Ray Projection}
\label{subsubsec:rigcraft_camera_ray_projection}
Given a single image $x$, its camera view $V$, and  $k$ 2D keypoints $\{\boldsymbol{l}_i\}_{i=1}^k (\boldsymbol{l}_{i}\in \mathbb{R}^2)$, each with confidence score $w_{i}$ detected by OpenPose \cite{Cao2019OpenPose}, we proceed to compute a 3D ray from camera center through 2D keypoint.
A camera view $V$ is defined by its intrinsic matrix $K$, rotation matrix $R$, and translation vector $\mathbf{t}$, which together determine how 3D points are projected into the image plane. To back-project a 2D keypoint $\boldsymbol{l}_j$ into 3D space, we first form its homogeneous coordinate $\tilde{\boldsymbol{l}}_j = \bigl[\boldsymbol{l}_j^\top,1\bigr]^\top$ and compute the corresponding ray direction and origin as follows:
\begin{equation}
  \mathbf{d}
  = \frac{R^\top K^{-1} \tilde{\boldsymbol{l}_j}}{\left\| R^\top K^{-1} \tilde{\boldsymbol{l}_j} \right\|_2},
  \quad
  \mathbf{o} = - R^\top \mathbf{t}.
\end{equation}
The resulting ray corresponding to keypoint $\boldsymbol{l}_j$ is $\mathbf{r} = \{\mathbf{o} + \lambda\,\mathbf{d}\mid \lambda\ge0\}$.

\subsubsection{Triangulation across camera viewpoints}
\label{subsubsec:rigcraft_triangulation}

To estimate the 3D position of a landmark \(\mathbf{\hat{l}}_i \in \mathbb{R}^3\), we use a set of rays \(\{\mathbf{r}^v\}_{v=1}^{m}\), each defined by an origin \(\mathbf{o}^v \in \mathbb{R}^3\) and a direction \(\mathbf{d}^v \in \mathbb{R}^3\), corresponding to \(m\) distinct camera views. In the absence of noise, these rays would ideally intersect at a single point. However, due to measurement noise and geometric uncertainty, exact intersection is typically not possible. We therefore seek the point \(\mathbf{\hat{l}}\) that minimizes the aggregate perpendicular distance to all rays. To compute these distances, we define the orthogonal projection matrix:
\begin{equation}
P(\mathbf d) \;=\; \mathbf{I} - \mathbf d\,\mathbf d^\top,
\end{equation}
which projects any vector onto the plane orthogonal to the ray direction \(\mathbf{d}^v\). For a candidate point \(\mathbf{l}\), the vector \(P(\mathbf{d}^v)(\mathbf{l} - \mathbf{o}^v)\) gives the component of \((\mathbf{l} - \mathbf{o}^v)\) orthogonal to the ray, and its norm corresponds to the shortest distance from \(\mathbf{l}\) to the ray \(\mathbf{r}^v\). We formulate triangulation as a \emph{weighted least-squares} problem:
\begin{equation}
\label{eq:rigcraft-triang}
\mathbf{\hat{l}}_i = \arg\min_{\mathbf{l}_i} \sum_{v=1}^{m}
\left(w^{v}_i \left\|P(\mathbf{d}^v)(\mathbf{l}_i - \mathbf{o}^v)\right\|_2\right)^2,
\end{equation}
where $w^v_i$ denotes the OpenPose confidence score for the
$i^{\mathrm{th}}$ keypoint in view $v$. This optimization is solved using a damped \emph{Levenberg--Marquardt} algorithm, initialized at the centroid of the camera origins \(\{\mathbf{o}^v\}_{v=1}^{m}\), which we empirically found to be more stable.

\subsubsection{Temporal Regularisation across multi-Frames}

To suppress pose jitter across frames and enforce smooth, plausible
motion, we stabilize each 3D landmark’s position by applying a discrete
Savitzky–Golay filter.
Given a particular frame and landmark, let $\{\mathbf{l}_j\}_{j=-M}^M$ be its
3D positions over a local window of $2M+1$ frames centered at that frame. The
smoothed position $\hat{\mathbf l}$ is
\begin{equation}
\label{eq:sg-single}
\hat{\mathbf l}
\;=\;
\sum_{j=-M}^{M} \beta_j \,\mathbf{l}_j,
\end{equation}
where the filter coefficients $\{\beta_j\}$ depend only on the window size
$M$ and polynomial degree $P$. They are obtained by locally fitting a
degree-$P$ polynomial
\begin{equation}
\label{eq:degree-p-polynomial}
p(j) \;=\; \sum_{n=0}^P \alpha_n\,j^n
\end{equation}
to the trajectory $\{\mathbf l_j\}$ in the least-squares sense over
$-M \le j \le M$:
\begin{equation}
\label{eq:sg-weights-single}
\alpha^*
\;=\;
\arg\min_{\alpha}
\sum_{j=-M}^{M}
\Bigl\|
  \mathbf{l}_{j}
  - p(j)
\Bigr\|_2^2,
\end{equation}
in which $n\in\{0,\dots,P\}$ is the polynomial degree and $\alpha_n$ are the corresponding least-squares–fitted coefficients for each power $j^n$.
The smoothed position at the center frame is the fitted value $p(0) = \alpha_0^*$, which can be written as a linear combination of the windowed positions as in~\eqref{eq:sg-single}.
Evaluating the normal equations of the associated Vandermonde system yields a closed-form expression for the filter coefficients $\{\beta_j\}$.
Repeating this per-frame smoothing across all frames produces the final temporally coherent 3D trajectories.
\emph{RigCraft} iteratively performs the above operation for each landmark in each image, producing 3D landmarks $\{\mathbf{l}_{i}\}_{i=1}^{k}$ for each frame in the sequence.

\subsection{PoseCraft: 3D Control \& Diffusion}
\label{subsec:method:posecraft}

\begin{figure}[t]
  \centering
  \includegraphics[width=0.95\linewidth]{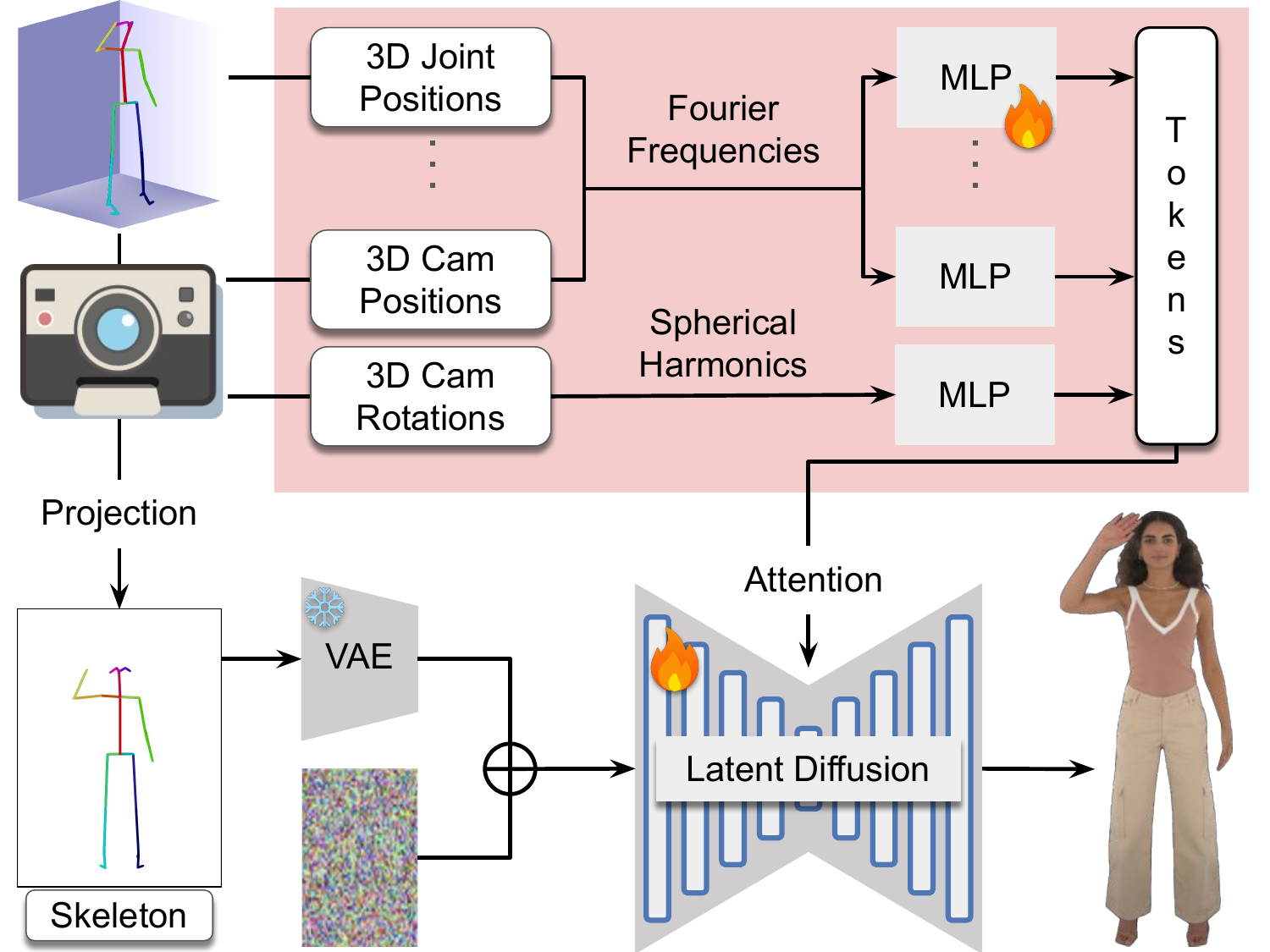}
  \caption{
  \textbf{PoseCraft Architecture.}
  We guide a latent diffusion UNet with two conditioning streams.
  First, a 3D Control Tokenizer converts camera extrinsics and 3D body landmarks into discrete tokens, which are injected via cross-attention for explicit 3D control.
  Second, a 2D skeleton projection is concatenated with the noisy latent input to provide direct spatial guidance.
  }
  \label{fig:posecraft_ov}
\end{figure}

\subsubsection{Preliminary}
The latent diffusion model~\cite{rombach2022ldm} comprises a variational autoencoder (VAE)~\cite{kingma2013auto} and a denoising UNet~\cite{ronneberger2015u}. 
The VAE consists of an encoder $\mathcal{E}$ that maps input image $x$ to a compact latent representation, $z = \mathcal{E}(x)$, and a decoder $\mathcal{D}$ that reconstructs the image from its latent representation, $x' = \mathcal{D}(z)$, enabling efficient image generation.
During training, latent features $z$ are corrupted over $t$ time steps to produce noisy features $\tilde{z}_t$.
Given time step $t$ and a task-specific condition $c$, a denoised UNet $\epsilon_{\theta}$ learns to predict the noise added to the noisy image $\tilde{z}_t$ with objective function:
 \begin{equation}
L_{DM} = \mathbb{E}_{z, \epsilon \in  \mathcal{N}(0,1)} [ || \epsilon - \epsilon_{\theta}(\tilde{z}_t, t, c)||^2_2 ].
\end{equation}
Finally, the predicted denoised latents $z'$ are decoded to recover the predicted image $x' = \mathcal{D}(x')$.

Our PoseCraft is built upon latent diffusion model with modified UNet to inject camera and pose condition, where the UNet is trained from scratch for each subject.
Following \cite{rombach2022ldm}, we finetune the \textit{decoder} of autoencoder $\{\mathcal{E},\mathcal{D}\}$ using perceptual loss and adversarial objective on per subject's training images.
This training strategy ensures that reconstructions lie in the per-subject distribution by enforcing local realism and is crucial for recovering sharper, more photorealistic details in the synthesized output images. 

\subsubsection{3D Control Tokeniser}
Given input image $x$ and its camera view $V$ and 3D landmarks ${\{\mathbf{l}_i\}_{i=1}^{k}}$, our \textit{3D Control Tokenizer} aims to inject view-aware and shape-aware information into denoised UNet. 
We assume the intrinsic parameters are fixed and incorporate extrinsic camera information via tailored encodings of both rotation and translation.

For camera rotation $\mathbf{a}$, we convert $R$ into Euler angles $\mathbf{a}=(\theta,\phi,\psi)$ and apply spherical harmonic encoding as follows:
\begin{equation}
  c(\mathbf{a})
  = [Y_{0}^{0}(\mathbf{a}),\,Y_{1}^{-1}(\mathbf{a}),\,Y_{1}^{0}(\mathbf{a}),\,Y_{1}^{1}(\mathbf{a}), \dots,Y_{d}^{d}(\mathbf{a})],
\end{equation}
where $Y$ denotes real spherical harmonic basis functions and \(d\) is maximum degree. 

For camera translation $\mathbf{t}$, we adopt positional encoding using a standard multi frequency embedding function:

\begin{equation}
\begin{split}
    c(\mathbf{t}) = [&\sin(2^0 \pi \mathbf{t}), \cos(2^0 \pi \mathbf{t}), \ldots, \\
            &\sin(2^{F-1} \pi \mathbf{t}), \cos(2^{F-1} \pi \mathbf{t})],
\end{split}
\end{equation}
where $F$ is the embedding depth. We use human body joints as pose guidance for diffusion model. Each joint position is treated as a 3D point and encoded using the same positional encoding scheme applied to the camera translation. For a joint position $\mathbf{l}_{i}$:
\begin{equation}
\begin{split}
    c(\mathbf{l}_{i}) = [&\sin(2^0 \pi \mathbf{l}_{i}), \cos(2^0 \pi \mathbf{l}_{i}), \ldots, \\
            &\sin(2^{F-1} \pi \mathbf{l}_{i}), \cos(2^{F-1} \pi \mathbf{l}_{i})].
\end{split}
\end{equation}

For each condition vector in $\{c(\mathbf{a}), c(\mathbf{t}), \{c(\mathbf{l}_{i})\}_{i=1}^{k}\}$, it is first processed through a dedicated multi-layer perceptron (MLP) to project it into a unified feature space. 
These projected embeddings $\{\mathbf{e}_{i}\}_{i=1}^{K+2}$ are then passed into cross-attention layers to guide the generation process. 

To further incorporate explicit landmark spatial configurations, we leverage 2D skeleton maps obtained by RigCraft projection.
These skeleton maps are encoded using $\mathcal{E}$, producing a latent pose representation $\mathbf{p}$.  
Then we concatenate the noisy latents $z_{t}$ with $\mathbf{p}$ as input to UNet. 
Architecturally, the input layer of the UNet accepts four additional channels for pose representations.
We learn the denoising UNet based on image-conditioning  pairs ($\tilde{z}_t$, Camera, Pose) via 
\begin{equation}
\begin{aligned}
    L_{DM} & = \mathbb{E}_{ z, \epsilon \in \mathcal{N}(0,1)} [\\& || \epsilon - \epsilon_{\theta}([\tilde{z}_t,\mathbf{p}], t, \{c(\mathbf{l}_{i})\}_{i=1}^{k},c(\mathbf{a}),c(\mathbf{t}))||^2_2 ].
\end{aligned}
\end{equation}

\begin{table}
  \centering
    \resizebox{0.9\linewidth}{!}{
  \begin{tabular}{c c c c c c c}
    \toprule
    Actor   & Seq.   & Gender   & Clothing     & Train     & Test      & Views \\
    \midrule
    \multirow{2}{*}{01}
      & 1 & \multirow{2}{*}{F} & \multirow{2}{*}{Loose} & 0--2000          & 2000--2200       & 100 \\
      & 2 &                          &                       & 0--2200          & --               & 100 \\
    \midrule
    \multirow{2}{*}{02}
      & 1 & \multirow{2}{*}{M}   & \multirow{2}{*}{Tight} & 0--1900          & 1900--2100       & 100 \\
      & 2 &                          &                       & 0--2400          & --               & 100 \\
    \midrule
    \multirow{2}{*}{06}
      & 1 & \multirow{2}{*}{F} & \multirow{2}{*}{Tight} & 0--2100          & 2100--2300       & 100 \\
      & 2 &                          &                       & 0--2400          & --               & 100 \\
    \bottomrule
  \end{tabular}}
  \vspace{-2mm}
  \caption{\textbf{Train/test split} for three selected \emph{ActorsHQ}~\cite{isik2023humanrf} subjects.
    For each actor and sequence, we list gender, clothing style, training-frame and held-out test-frame ranges, and the number of virtual camera views per frame.
    We specifically choose Actor 01 (female, loose clothing), Actor 02 (male, tight clothing), and Actor 06 (female, tight clothing) to cover both genders and garment variety for comprehensive validation coverage.
  }

  \label{tab:train-test-split}
\end{table}

\section{Experiments}
\label{sec:experiments}

\begin{figure}
  \centering
  \includegraphics[width=0.95\linewidth]{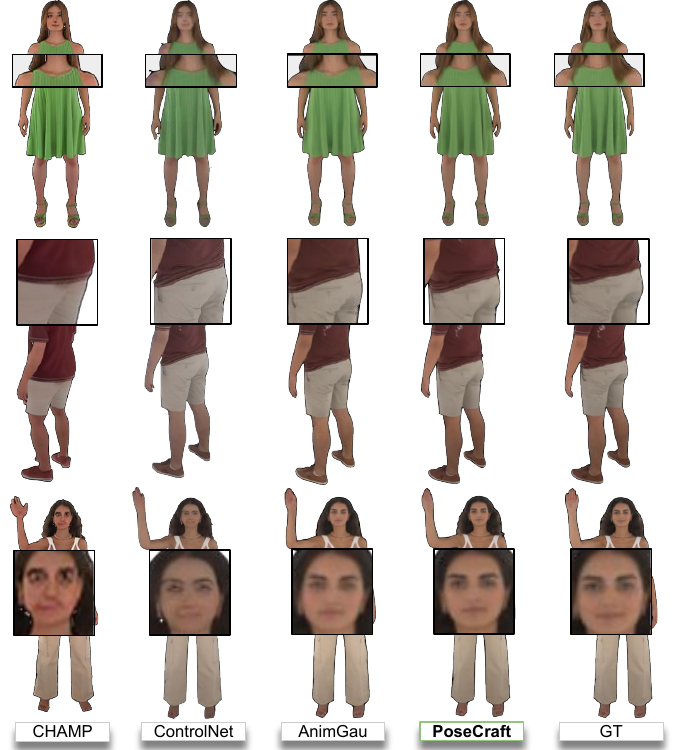}
  \caption{\textbf{Qualitative comparison} on the \emph{GenHumanRF} test split~\cite{isik2023humanrf}. 
  The proposed PoseCraft method largely outperforms the image-based
  CFLD~\cite{lu2024cfld}, CHAMP~\cite{zhu2024champ}, T2I-Adapter~\cite{mou2024t2i} and ControlNet~\cite{zhang2023controlnet}, and compare better or comparatively to the volumetric rendering SOTA Animatable Gaussian~\cite{li2024animatablegaussians}.
  }
  \label{fig:qualitative}
\end{figure}

\begin{table*}
  \centering
  \resizebox{0.9\linewidth}{!}
{
  \begin{tabular}{l 
                  *{4}{c}  %
                  *{4}{c}  %
                  *{4}{c}  %
                 }
    \toprule
    \multirow{2}{*}{Method}               &
    \multicolumn{4}{c}{\textsc{Actor01}} &
    \multicolumn{4}{c}{\textsc{Actor02}} &
    \multicolumn{4}{c}{\textsc{Actor06}} \\
    \cmidrule(lr){2-5}   \cmidrule(lr){6-9}   \cmidrule(lr){10-13}
                         &
    PSNR$\uparrow$       & SSIM$\uparrow$      & LPIPS$\downarrow$   & FID$\downarrow$     &
    PSNR$\uparrow$       & SSIM$\uparrow$      & LPIPS$\downarrow$   & FID$\downarrow$     &
    PSNR$\uparrow$       & SSIM$\uparrow$      & LPIPS$\downarrow$   & FID$\downarrow$     \\
    \midrule
    CFLD~\cite{lu2024cfld} &
    16.01                & 0.8035              & 0.2508              & 120.49              &
    15.91                & 0.7843              & 0.2823              & 116.08              &
    16.15                & 0.8149              & 0.2270              & 101.85              \\
    AnimateAnyone~\cite{hu2024animate} &
    19.70               & 0.8438              & 0.1508              & 70.27              &
    20.47                & 0.8398              & 0.1591              & 59.55             &
    18.13                & 0.8299              & 0.1630              & 104.82              \\
    CHAMP~\cite{zhu2024champ} &
    20.75               & 0.8536              & 0.1270              & 53.67              &
    21.12                & 0.8559              & 0.1352              & 42.00              &
    19.04                & 0.8601              & 0.1418              & 66.94              \\
    T2I-Adapter~\cite{mou2024t2i} &
    23.01                & 0.9042              & 0.1053              & 41.89               &
    25.16                & 0.9217              & 0.0823              & 38.32               &
    23.99                & 0.9268              & 0.0798              & 40.34               \\
    ControlNet~\cite{zhang2023controlnet} &
    26.93                & 0.6737              & 0.0489              & 34.72               &
    23.02                & 0.2080              & 0.0944              & 27.65               &
    26.66                & 0.4622              & 0.0494              & 29.08               \\
    Animatable Gaussians~\cite{li2024animatablegaussians} &
    27.90                & \cellcolor{LimeGreen!50}0.9437     & 0.0428              & \cellcolor{LimeGreen!50}21.42      &
    \cellcolor{LimeGreen!50}30.11       & 0.9570              & \cellcolor{LimeGreen!50}0.0289     & \cellcolor{LimeGreen!50}15.53     &
    28.40                & 0.9562              & \cellcolor{LimeGreen!50}0.0338     & 23.96              \\
    \midrule    
    \textbf{PoseCraft} &
    \cellcolor{LimeGreen!50}28.11      & \cellcolor{YellowGreen!30}0.9380              & \cellcolor{LimeGreen!50}0.0420     & \cellcolor{YellowGreen!30}24.53               &
    \cellcolor{YellowGreen!30}30.01                & \cellcolor{LimeGreen!50}0.9577    & \cellcolor{YellowGreen!30}0.0322              & \cellcolor{YellowGreen!30}21.83               &
    \cellcolor{LimeGreen!50}28.69       & \cellcolor{LimeGreen!50}0.9584     & \cellcolor{YellowGreen!30}0.0350              & \cellcolor{LimeGreen!50}23.39      \\
    \bottomrule
  \end{tabular}
}
\vspace{-2mm}
    \caption{\textbf{Quantitative comparison} of methods on our \emph{GenHumanRF} train/test split.
    Our PoseCraft method largely outperforms diffusion-based CFLD~\cite{lu2024cfld}, AnimateAnyone~\cite{hu2024animate}, CHAMP~\cite{zhu2024champ}, T2I-Adapter~\cite{mou2024t2i}, ControlNet~\cite{zhang2023controlnet}, and compares better or competitively to best volumetric rendering SOTA Animatable Gaussians~\cite{li2024animatablegaussians}.
    ($\uparrow$ higher is better, $\downarrow$ lower is better, boldface indicates best result.)
  }
  \label{tab:quantitative-comparison}
\end{table*}

\subsection{GenHumanRF: Dataset Generation}
\label{subsec:experiments:genhumanrf}

We implement \emph{GenHumanRF} dataset generation pipeline from the multi-camera capture dataset \textit{ActorsHQ}  and the neural rendering framework \textit{HumanRF}~\cite{isik2023humanrf}. ActorsHQ captures narrow-FOV images focusing on  body parts from several subjects, tailored for Neural Rendering such as HumanRF. Our \emph{GenHumanRF} pipeline is designed to render full-body images suitable for diffusion learning.

We partition each 2,200-frame sequence from ActorsHQ~\cite{isik2023humanrf} into shorter intervals, train separate HumanRF models per interval, and then render every frame from 100 virtual cameras evenly spaced in a horizontal circle. This yields around 440,000 full-body images per actor (512$\times$384 resolution), each paired with its virtual camera’s calibration. For evaluation we focus on Actors 01, 02, and 06 to cover both genders and loose vs. tight clothing variations.

\subsection{Experiment Setup}
\label{subsec:exp-setup}

\noindent\textbf{Dataset split.}
All experiments use our \emph{GenHumanRF} dataset pipeline, designed specifically for diffusion-based training.
We adopt the training/testing partitions summarized in \cref{tab:train-test-split}, selecting three subjects (Actors 01, 02, and 06) to cover both genders and loose vs.\ tight clothing.
For each actor and sequence, \cref{tab:train-test-split} lists the frame ranges reserved for training and held-out testing, with 100 virtual camera views per frame.
This ensures consistent evaluation across varied appearance and motion conditions.

\noindent\textbf{Evaluation.}
We compare \emph{PoseCraft} with several state-of-the-art methods, including
CFLD~\cite{lu2024cfld},
AnimateAnyone~\cite{hu2024animate}
CHAMP~\cite{zhu2024champ},
T2I-Adapter~\cite{mou2024t2i},
ControlNet~\cite{zhang2023controlnet},
and Animatable Gaussians~\cite{li2024animatablegaussians}. 
T2I-Adapter~\cite{mou2024t2i}, ControlNet~\cite{zhang2023controlnet}, and Animatable Gaussians~\cite{li2024animatablegaussians} are re‑trained on our split using official implementations.
We evaluate image fidelity and perceptual realism using PSNR, SSIM \cite{wang2004image}, LPIPS \cite{zhang2018lpips}, and FID \cite{heusel2017gans}.

\noindent\textbf{Implementation details.} 
We implement our method using PyTorch Lightning \cite{Falcon2019PyTorchLightning} and train on a single A100 (80 GB) GPU with 16-bit mixed precision.
We use AdamW optimizer \cite{loshchilov2019adam_optimizer} with learning rate $10^{-4}$ and weight decay $10^{-5}$, a constant learning rate schedule with 2k warmup steps, and train for 200k steps with batch size 32. Images are resized to $512 \times 384$ resolution.
The UNet is trained from scratch and configured
with 8 input channels and 4 output channels, where the input concatenates the noisy image latent with the pose latent.
Each layer has 2 blocks and the out-channels are set to [64, 128, 256, 256], and cross-attention dimension is 768 with 8 attention heads.
Within 3D Control Tokenizer, spherical harmonics is used for camera rotation and Fourier features is used for all translation.
The stage-2 processing employs 27 MLPs that processes 27 tokens (1 rotation, 1 translation, 25 joint tokens), where each token's MLP maps from 192 to 128 to 768 dimensions.
The total number of trainable parameters (UNet + 3D Control Tokenizer) is approximately 33.3M.

\subsection{Qualitative Comparison}
\label{subsec:qualitative}

\Cref{fig:qualitative} presents a qualitative comparison of PoseCraft against state-of-the-art baselines.
Across a variety of challenging poses and viewpoints, PoseCraft consistently produces sharp, coherent silhouettes and preserves fine high-frequency details (\eg folds, hair strands, and embroidery) that are either blurred or distorted by purely 2D image–based methods  \cite{lu2024cfld, zhang2023controlnet}. More qualitative comparisons are provided in supplementary.

Compared to the volumetric-based SOTA Animatable Gaussians~\cite{li2024animatablegaussians} which also deliver high‐fidelity renderings, PoseCraft achieves improved or comparable perceptual results-while requiring no per-scene template re-optimization or dense point‐cloud fitting.
This demonstrates that our 2D latent diffusion conditioned on 3D Control Tokenizer can match the strengths of recent 3D representations at reduced computational and modelling complexity.

\subsection{Quantitative Comparison}
\label{subsec:quantitative}

\Cref{tab:quantitative-comparison} reports 
PSNR, SSIM \cite{wang2004image}, LPIPS \cite{zhang2018lpips}, and FID \cite{heusel2017gans}
on three actors from the \emph{GenHumanRF} test split.
Across all actors, PoseCraft substantially outperforms the 2D image-based baselines (CFLD~\cite{lu2024cfld} and ControlNet~\cite{zhang2023controlnet}), achieving gains of 10-12 dB in PSNR, 0.15-0.20 in SSIM, and reducing LPIPS and FID by over 50\% in most cases.  

Against the volumetric Animatable Gaussians \cite{li2024animatablegaussians}, PoseCraft matches or slightly exceeds performance on key perceptual metrics: it achieves the highest PSNR on Actor01 and Actor06, the best SSIM on Actor02 and Actor06, and the lowest LPIPS on Actor01, while maintaining FIDs within 2-3 points of the best.
These results demonstrate that our 2D diffusion-based pipeline can rival state-of-the-art 3D reconstruction approaches, combining superior image fidelity with reduced modeling complexity.

\begin{table}
  \centering
  \resizebox{0.95\linewidth}{!}
  {
  \begin{tabular}{l l c c c c}
    \toprule
      Camera \& Skeleton & Test & PSNR$\uparrow$ & SSIM$\uparrow$ & LPIPS$\downarrow$ & FID$\downarrow$ \\
      \midrule
      \multirow{3}{*}{Plucker \& OP-map}
        & view           & 26.28 & 0.9505 & 0.0385 & \cellcolor{LimeGreen!50}{20.83} \\
        & frame          & 29.05 & 0.9620 & 0.0278 & 30.78 \\
        & all            & 26.23 & 0.9518 & 0.0398 & 22.60 \\
      \midrule
      \multirow{3}{*}{Plucker \& RC-map}
        & view           & 27.16 & 0.9562 & 0.0359 & 22.32 \\
        & frame          & 30.32 & 0.9679 & 0.0247 & 31.87 \\
        & all            & 27.49 & 0.9580 & 0.0360 & 23.59 \\
      \midrule
      \multirow{3}{*}{\textbf{SH} \& RC-map}
        & view           & \cellcolor{LimeGreen!50}{28.10} & \cellcolor{LimeGreen!50}{0.9557} & \cellcolor{LimeGreen!50}{0.0355} & \cellcolor{YellowGreen!30}21.20 \\
        & frame          & \cellcolor{LimeGreen!50}{30.66} & \cellcolor{LimeGreen!50}{0.9692} & \cellcolor{LimeGreen!50}{0.0229} & \cellcolor{LimeGreen!50}{29.39} \\
        & all            & \cellcolor{LimeGreen!50}{28.69} & \cellcolor{LimeGreen!50}{0.9595} & \cellcolor{LimeGreen!50}{0.0328} & \cellcolor{LimeGreen!50}{23.39}  \\
    \bottomrule
  \end{tabular}
  }
  \vspace{-2mm}\caption{\textbf{Camera encoding ablation}.
    Spherical Harmonics (SH) for camera rotation achieves better metrics than Plucker Ray with either RigCraft skeleton or OpenPose skeleton.
  }
  \label{tab:camera-input-comparison}
\end{table}

\begin{table}
  \centering
  \resizebox{0.95\linewidth}{!}
  {
  \begin{tabular}{l l c c c c}
    \toprule
    Token Embedding & Test & PSNR$\uparrow$ & SSIM$\uparrow$ & LPIPS$\downarrow$ & FID$\downarrow$ \\
    \midrule
    \multirow{3}{*}{No MLP}
      & view           & 27.50 & 0.9500 & 0.0386 & 21.94 \\
      & frame          & 29.53 & 0.9627 & 0.0255 & 31.08 \\
      & all           & 28.10 & 0.9532 & 0.0376 & 23.57 \\
    \midrule
    \multirow{3}{*}{Share MLP}
      & view           & 28.09 & 0.9534 & 0.0362 & 23.39 \\
      & frame          & 29.27 & 0.9616 & 0.0264 & 31.88 \\
      & all           & 28.37 & 0.9548 & 0.0360 & 24.33 \\
    \midrule
    \multirow{3}{*}{\textbf{Per-token MLP}}
      & view           & \cellcolor{LimeGreen!50}{28.10} & \cellcolor{LimeGreen!50}{0.9557} & \cellcolor{LimeGreen!50}{0.0355} & \cellcolor{LimeGreen!50}{21.20} \\
      & frame          & \cellcolor{LimeGreen!50}{30.66} & \cellcolor{LimeGreen!50}{0.9692} & \cellcolor{LimeGreen!50}{0.0229} & \cellcolor{LimeGreen!50}{29.39} \\
      & all           & \cellcolor{LimeGreen!50}{28.69} & \cellcolor{LimeGreen!50}{0.9595} & \cellcolor{LimeGreen!50}{0.0328} & \cellcolor{LimeGreen!50}{23.39}  \\
    \bottomrule
  \end{tabular}
  }
  \vspace{-2mm}
  \caption{\textbf{3D Control Tokenizer ablation.} 
  Our \emph{Per-token MLP} treats each 3D information token independently to elevate model's 3D awareness, outperforming \emph{No MLP} and \emph{Shared MLP} ablations.
  }
  \vspace{-2mm}
  \label{tab:ablation-token-embedding}
\end{table}

\subsection{Ablation Study}
\label{sec:ablation}

We perform ablation study on Actor06
to isolate the impact of key components in our PoseCraft architecture.

\noindent\textbf{Camera Rotation Projection}:
\cref{tab:camera-input-comparison} compares camera rotation projection strategy
between \emph{Plucker Ray} and our \emph{Spherical Harmonics}.
Our Spherical Harmonics camera rotation projection outperforms Plucker Ray on both \emph{RigCraft} and \emph{OpenPose} skeletons.

\noindent\textbf{Token Embedding Strategy}:
we compare in \cref{tab:ablation-token-embedding} three strategies for embedding projected 3D inputs of camera rotation, translation, and landmark coordinates:
(1) \emph{No MLP} uses direct zero-padded token vectors;
(2) \emph{Shared MLP} applies a single MLP to all tokens; and
(3) our \emph{Per-token MLP} uses a separate MLP for each token type.
Our Per-token MLP outperforms both ablations, demonstrating the value of dedicated, type-specific embeddings.

\noindent\textbf{Diffusion Control Method}:
\Cref{tab:ablation-control} evaluates combinations of control signals to condition the latent diffusion model:
(1) only using our {3D Control Tokenizer} (3D-CT);
(2) only using OpenPose skeleton (OP-Map);
(3) using OpenPose skeleton with our {3D-CT};
(4) only using our \emph{RigCraft} skeleton;
(5) using our \emph{RigCraft} skeleton with our {3D-CT}
Replacing OpenPose skeleton with RigCraft skeleton consistently improves fidelity and realism, and fusing RigCraft skeleton with our 3D Control Tokenizer yields the best scores across PSNR, SSIM, LPIPS, and FID, underscoring their complementary strengths.

\begin{table}
  \centering
  \resizebox{0.95\linewidth}{!}
  {
  \begin{tabular}{l l c c c c}
    \toprule
    Control Method & Test & PSNR$\uparrow$ & SSIM$\uparrow$ & LPIPS$\downarrow$ & FID$\downarrow$ \\
    \midrule
    \multirow{3}{*}{3D-CT \emph{only}}
      & view   & 24.59 & 0.9526 & 0.05883 & 28.25 \\
      & frame  & 28.26 & 0.9574 & 0.02858 & 32.19 \\
      & all    & 25.45 & 0.9471 & 0.0553  & 28.32 \\
    \midrule
    \multirow{3}{*}{OP-Map \textit{only}}
      & view   & 26.73 & 0.9443 & 0.0427 & 24.04 \\
      & frame  & 28.56 & 0.9570 & 0.0302 & 34.53 \\
      & all    & 26.95 & 0.9456 & 0.0433 & 25.05 \\
    \midrule
    \multirow{3}{*}{OP-Map \& 3D-CT}
      & view   & 27.38 & 0.9495 & 0.0391 & 21.65  \\
      & frame  & 29.55 & 0.9633 & 0.0250 & 30.37 \\
      & all    & 27.66 & 0.9515 & 0.0394 & 23.40  \\
    \midrule
    \multirow{3}{*}{RC-Map \textit{only}}
      & view   & 27.46 & 0.9507 & 0.0379 & 23.29 \\
      & frame  & 30.67 & 0.9674 & 0.0233 & 31.39 \\
      & all    & 28.31 & 0.9545 & 0.0368 & 24.65 \\
    \midrule
    \multirow{3}{*}{\textbf{RC-Map} \& \textbf{3D-CT}}
      & view   & \cellcolor{LimeGreen!50}{28.10} & \cellcolor{LimeGreen!50}{0.9557} & \cellcolor{LimeGreen!50}{0.0355} & \cellcolor{LimeGreen!50}{21.20} \\
      & frame  & \cellcolor{LimeGreen!50}{30.66} & \cellcolor{LimeGreen!50}{0.9692} & \cellcolor{LimeGreen!50}{0.0229} & \cellcolor{LimeGreen!50}{29.39} \\
      & all    & \cellcolor{LimeGreen!50}{28.69} & \cellcolor{LimeGreen!50}{0.9595} & \cellcolor{LimeGreen!50}{0.0328} & \cellcolor{LimeGreen!50}{23.39} \\
    \bottomrule
  \end{tabular}
  }
  \vspace{-2mm}
    \caption{\textbf{Control method ablation.} 
    Our \emph{RigCraft} (RC) skeleton together with \emph{3D Control Tokenizer} (3D-CT) architecture outperforms all other control combinations, including
    \emph{3D-CT only};
    \emph{OP(OpenPose)-map only};
    \emph{OP-Map \& 3D-CT}; and
    \emph{RC-Map only}.
    }
  \label{tab:ablation-control}
\end{table}

\begin{figure}[t]
  \centering
  \includegraphics[width=0.88\linewidth]{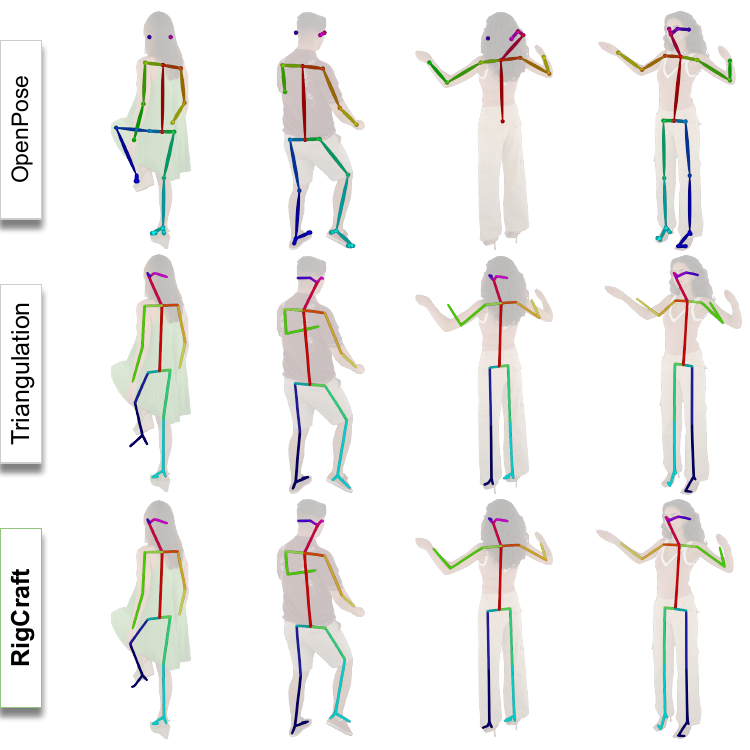}
  \vspace{-2mm}
  \caption{Qualitative visualization of the \textit{RigCraft} 3D landmark estimation and refinement. From top to bottom:  (1) noisy 3D landmarks from re-projected 2D OpenPose detections; (2) landmarks after multi-view triangulation; and (3) the final, temporally coherent RigCraft outputs after smoothing.
  }
  \label{fig:rigcraft-qualitative-comparison-white-bg}
\end{figure}

\section{Discussion}

A central contribution of PoseCraft is the use of discrete 3D tokens
as the primary conditioning signal for our diffusion backbone.
By moving beyond 2D heatmaps or skeleton maps alone, these tokens elevate the network with explicit geometric priors that (a) preserve identity and image quality, (b) disentangle pose from appearance, and (c) enable precise control over novel viewpoints.
Empirically, this design yields sharp silhouettes, coherent limb articulations, and faithful recovery of high-frequency details (\eg garment folds, hair strands) under challenging poses, matching or exceeding volumetric baselines without costly template fitting.
Despite its strengths, PoseCraft still exhibits several practical limitations:
\noindent\textbf{Single-identity training}: our current pipeline learns one model per actor, preventing zero-shot synthesis of unseen identities. Extending to identity-agnostic or few-shot adaptation (\eg via a learned identity token-remains an open challenge).
\noindent\textbf{Garment topology}: while 3-D landmarks robustly capture articulated geometry, they struggle to represent loose or multi-layered garments (\eg skirts, scarves), occasionally yielding “ghost” limbs or texture bleeding.
\noindent\textbf{Hand articulation}: we currently omit detailed hand keypoints, treating the wrist as a single joint.
Incorporating a dedicated hand encoding branch would help resolve finger-level articulation and improve realism.

\section{Conclusion}
\label{sec:conclusion}
We introduce \textbf{PoseCraft}, a diffusion framework that exposes explicit 3D control by feeding sparse 3D body landmarks and camera extrinsics to the denoiser as discrete tokens, and \textbf{RigCraft}, a lightweight multi-view triangulation module that turns 2D detections into temporally stable 3D landmarks.
This tokenized 3D interface differs fundamentally from prior 2D or 3D-controller image synthesis that relies on rasterized 2D maps or parametric templates.
By incorporating direct 3D conditioning, PoseCraft avoids re-projections ambiguities and improve viewpoint consistency while preserving identity and high-frequency details without mesh rigging or template optimisation.
Coupled with \textbf{GenHumanRF} for data synthesis, our experiments show substantial gains over 2D pose-guided diffusions and competitive results than best volumetric renderer while retaining simpler pipeline.
Ablations further confirm the value of our \emph{3D Control Tokenizer} within PoseCraft architecture and the effectiveness of our RigCraft landmark estimations.

{
    \small
    \bibliographystyle{ieeenat_fullname}
    \bibliography{reference}
}

\end{document}